\documentclass{article}

\usepackage[preprint]{nips_2018}

\usepackage[utf8]{inputenc} 
\usepackage[T1]{fontenc}    
\usepackage{url}            
\usepackage{booktabs}       
\usepackage{amsfonts}       
\usepackage{nicefrac}       
\usepackage{tikz}
\usepackage{microtype}      

\usepackage{amssymb,amsmath,amsthm}
\usepackage{bm}
\usepackage{mathtools}
\usepackage{dsfont}

\def\K{\bm{K}}
\def\V{\bm{V}}
\def\W{\bm{W}}

\def\a{\bm{a}}
\def\b{\bm{b}}

\def\q{\bm{q}}

\def\s{\bm{s}}

\def\u{\bm{u}}
\def\x{\bm{x}}
\def\y{\bm{y}}

\def\R{\mathds{R}}

\def\one{\mathds{1}}
\def\zero{\bm{0}}
\def\opt{\star}
\DeclareMathOperator*\argmin{argmin}
\newcommand{\htanh}{\mathrm{hardtanh}}
\newcommand{\relu}{\mathrm{ReLU}}
\newcommand{\prox}{\mathrm{prox}}
\newcommand{\norm}[1]{\Vert#1\Vert}

\numberwithin{equation}{section}

\title{Primal-dual residual networks}

\author{
  Christoph Brauer\\
  Institute of Analysis and Algebra\\
  TU Braunschweig\\
  Braunschweig, Germany \\
  \texttt{ch.brauer@tu-braunschweig.de} \\
  \And
  Dirk Lorenz\\
  Institute of Analysis and Algebra\\
  TU Braunschweig\\
  Braunschweig, Germany\\
  \texttt{d.lorenz@tu-braunschweig.de}
}

\begin{document}

\maketitle

\begin{abstract}
  In this work, we propose a deep neural network architecture motivated by primal-dual splitting methods from convex optimization. We show theoretically that there exists a close relation between the derived architecture and residual networks, and further investigate this connection in numerical experiments. Moreover, we demonstrate how our approach can be used to unroll optimization algorithms for certain problems with hard constraints. Using the example of speech dequantization, we show that our method can outperform classical splitting methods when both are applied to the same task.
\end{abstract}

\section{Introduction}
\label{sec:introduction}

The task to recover multivariate continuous data from noisy indirect measurements arises in many applications in signal and image processing and beyond. Here, we consider the typical situation where an unknown ground truth signal $\x^{\dagger}\in \R^{n}$ shall be reconstructed from observations $\u\in\R^{m}$. In many cases of interest, the ground truth and the observations are connected via a linear model $\u = \K\x^{\dagger} + \bm{\eta}$. Therein, $\K\in\R^{m\times n}$ is a known measurement matrix and $\bm\eta\in\R^{n}$ is a random noise vector. In recent years, variational approaches in the form of
\begin{equation}
  \label{eq:primal_problem}
  \hat \x \in \argmin_{\x\in\R^{n}} \ F(\K\x) + G(\x)
\end{equation}
have been widely used for such reconstruction tasks. The function $F$ is usually chosen in accordance with the noise distribution and such that discrepancies between $\K\x$ and $\u$ are penalized appropriately. Moreover, the function $G$ incorporates prior knowledge about the ground truth and penalizes solutions which do not conform with these requirements.

In the following, we show how a certain class of algorithms designed to solve (\ref{eq:primal_problem}), namely proximal splitting methods, can be unrolled in terms of a neural network and discuss the resemblance of the resulting architecture to residual networks. From this comparison, we derive a new architecture combining the advantages of both approaches. To that end, we briefly review proximal splitting methods in Subsection \ref{sec:proximal_splitting_methods}, followed by an overview of related work in Subsection \ref{sec:related_work}. Further, Section \ref{sec:deep_pd_network} contains the derivation of our architecture and Section \ref{sec:speech_dequantization} describes its application to speech dequantization. In Section \ref{sec:numerical_experiments}, we report the results of our numerical experiments, before we finally conclude our work in Section \ref{sec:conclusion}.

\subsection{Proximal splitting methods}
\label{sec:proximal_splitting_methods}

There exist a variety of optimization methods to solve (\ref{eq:primal_problem}) and which one to choose depends particularly on the structure of the involved functions $F$ and $G$. If both are differentiable, then one could, for example, use any form of gradient descent. However, this is often not the case and one has to resort to alternative methods such as proximal splitting methods (see, e.g., \citet{parikh2013proximal} and~\citet{bauschke2017convex}). Among these, the algorithm proposed by \citet{Chambolle2011} has gained a lot of attention and turned out to perform well on numerous practical tasks. The algorithm is designed for convex optimization problems (\ref{eq:primal_problem}) with $F:\R^{m}\rightarrow\R\cup\{\infty\}$ and $G:\R^{n}\rightarrow\R\cup\{\infty\}$ being proper, convex and lower-semicontinuous functions. One iteration of the Chambolle--Pock algorithm consists of one dual and one primal update step, followed by a primal extrapolation step, i.e., 
\begin{equation}
  \label{eq:chambolle_pock}
  \begin{cases}
    \y^{k+1} = \prox_{\sigma F^{*}}(\y^{k} + \sigma\K\bar \x^{k}),\\
    \x^{k+1} = \prox_{\tau G^{\phantom{*}}}(\x^{k} - \tau\K^{\top}\y^{k+1}),\\
    \bar \x^{k+1} = \x^{k+1} + \theta(\x^{k+1} - \x^{k}).
  \end{cases}
\end{equation}
The step sizes $\sigma, \tau > 0$ need to be chosen subject to the operator norm of $\K$ and $\theta\in [0, 1]$ is an extrapolation factor. Furthermore, the mappings in the first two rows of (\ref{eq:chambolle_pock}) are the proximal operators of $\sigma F^{*}$ and $\tau G$, where $F^{*}$ is the convex conjugate function of $F$ (cf.~\citet{bauschke2017convex}). Many convex functions of interest have the property that the respective proximal mappings have simple closed-form representations which is crucial for the applicability of the Chambolle-Pock algorithm.

\subsection{Related work}
\label{sec:related_work}

The idea to unroll iterative methods for optimization problems and treat them as deep neural networks appeared previously in the literature. For example, \citet{Gregor2010} interpreted the iterative shrinkage-thresholding algorithm (ISTA) from sparse coding as a recurrent neural network and trained the involved linear operator as well as the shrinkage function to obtain learned ISTA (LISTA). More recently, \citet{Wang2016} proposed to unroll iterations (\ref{eq:chambolle_pock}) of the Chambolle-Pock algorithm in the context of proximal deep structured networks. Similarly, \citet{Riegler2016} and \citet{Riegler2016a} introduced deep primal-dual networks for depth super-resolution, where unrolled iterations are stacked on top of a deep convolutional network. In both cases, linear operators as well as step sizes and extrapolation factors are learned. In contrast, the learned primal-dual algorithm proposed by \citet{Adler2018} replaces the proximal operators in (\ref{eq:chambolle_pock}) by convolutional neural networks, and keeps the linear operator fixed.

\subsection{Our contribution}
\label{sec:our_contribution}

In this work, we propose to unroll a fixed number of Chambolle-Pock iterations (\ref{eq:chambolle_pock}) in a deep neural network. Among the above-mentioned approaches, ours is most closely related to \citep{Wang2016}. However, there are some important differences. First, we interpret the unrolled algorithm as a special kind of residual network \citep{He2016} rather than as a recurrent neural network. Based on this interpretation, we show that a minimal adjustment of the unrolled network architecture is enough to obtain a plain residual network, which motivates the term \emph{primal-dual residual network}. Second, we consider the common special case of (\ref{eq:primal_problem}) where $F$ is a norm and $G$ is the indicator function of a norm ball, i.e., where both functions encode a convex optimization problem with hard constraints. We deduce that in this case, the proximal operators of $F$ and $G^{*}$ are projections and further, that the step sizes in (\ref{eq:chambolle_pock}) can be learned implicitly through the involved linear operators. In the third place, we take up the application of speech dequantization. This application was addressed previously by \cite{Brauer2016} who also proposed to perform a limited number of Chambolle-Pock iterations, but in the classical sense with an a priori fixed linear operator.

\section{Network architecture}
\label{sec:deep_pd_network}

\subsection{Primal-dual networks}
\label{sec:primal_dual_networks}

\begin{figure}[t]
  \centering
  \begin{tikzpicture}[yscale= 2.4,xscale=2.4,
    place/.style={circle,
      draw=blue!50,
      fill=blue!20,thick,
      minimum width={1.1cm},},
    transition/.style={rectangle,draw=black!50,fill=black!20,thick}]]

    \node (yl) at (0,0) [place] {$\y^{[l]}$};

    \node (xl) at (1.5,1) [place] {$\x^{[l]}$};
    
    \node (xl+1) at (4.5,1) [place] {$\x^{[l+1]}$}; 
    
    \node (+yl+1) at (1.5,0) [transition] {$+$};
    
    \node (+xl+1) at (3,1) [transition] {$+$};+
    
    \node (yl+1) at (3,0) [place] {$\y^{[l+1]}$};

    \draw[->] (yl) to (+yl+1);
    
    \draw[->] (xl) to (+xl+1);
    
    \draw[->] (xl) to node [right,midway] {$\W^{[l+1]}$} (+yl+1);
    
    \draw[->] (+yl+1) to node [above,midway] {$\prox_{\sigma F^{*}}$} (yl+1);
    
    \draw[->] (yl+1) to node [right,midway] {$\V^{[l+1]}$} (+xl+1);
    
    \draw[->] (+xl+1) to node [above,midway] {$\prox_{\tau G}$} (xl+1);
    
    \draw[<-] (xl) -- ++(-0.325,0) edge[-,dashed] ++ (-0.325,0);
    
    \draw[<-] (yl) -- ++(-0.325,0) edge[-,dashed] ++ (-0.325,0);
    \draw (yl) -- ++(0,0.325) edge[->,dashed] ++ (0,0.325);

    \draw (xl+1) -- ++(0.325,0) edge[->, dashed] ++ (0.325,0);
    \draw (xl+1) -- ++(0,-0.325) edge[->, dashed] ++ (0,-0.325);
    
    \draw (yl+1) -- ++(0.325,0) edge[->, dashed] ++(0.325,0);
    
  \end{tikzpicture}
  \caption{Primal-dual block without extrapolation.}
  \label{fig:primal_dual_block}
\end{figure}
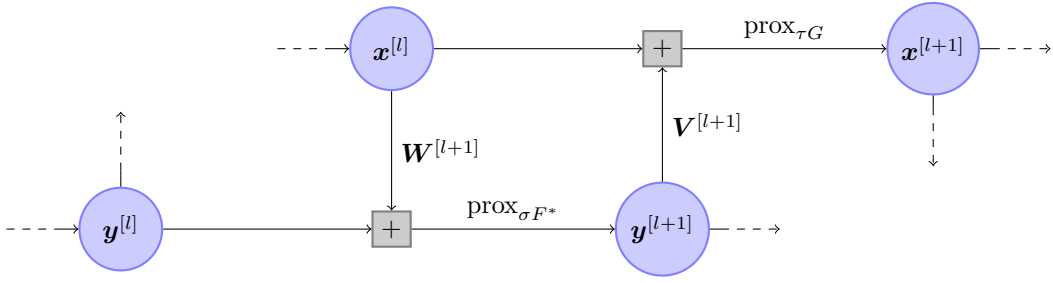

We first take the point of view that iterations (\ref{eq:chambolle_pock}) can be considered building blocks of feedforward neural networks. This conception makes sense in two respects: First, the proximal operators $\prox_{\sigma F^{*}}$ and $\prox_{\tau G}$ can be interpreted as activation functions. Second, the linear operator and its transpose can be considered manufactured weights which could just as well be learned. To that end, we regard $\y^{k+1} = \y^{[l+1]}$ and $\x^{k+1} = \x^{[l+1]}$ as activations, and $\sigma\K = \W^{[l+1]}$ as well as $\tau\K^{\top} = \V^{[l+1]}$ as associated weights. This notion leads us to \emph{primal-dual blocks} in the form of
\begin{equation}
  \label{eq:network_block}
  \begin{cases}
    \y^{[l + 1]} &\hspace{-.75em} = \prox_{\sigma F^{*}}(\W^{[l+1]}\bar\x^{[l]}+ \y^{[l]}),\\
    \x^{[l+1]} &\hspace{-.75em} = \prox_{\tau G}(\V^{[l+1]}\y^{[l+1]} + \x^{[l]}),\\
    \bar\x^{[l + 1]} &\hspace{-.75em} = \x^{[l+1]} + \theta(\x^{[l+1]} - \x^{[l]}).
  \end{cases}
\end{equation}
Hence, performing $L$ iterations (\ref{eq:chambolle_pock}) corresponds to stacking just as many primal-dual blocks with $\W^{[1]} = \dots = \W^{[L]} = \K$ and $\V^{[1]} = \dots = \V^{[L]} = \K^{\top}$. As opposed to this, our approach is to learn all weights and allow for differing weights in different blocks. In particular, we do not require that $(\W^{[l]})^{\top} = \V^{[l]}$. For simplicity, we restrict ourselves to the case $\theta = 0$ in the following. In other words, we omit the extrapolation in (\ref{eq:network_block}) and take $\bar\x^{[l+1]} = \x^{[l+1]}$ throughout. Figure~\ref{fig:primal_dual_block} illustrates a primal-dual block without extrapolation. Therein, $\y^{[l]}$ and $\x^{[l]}$ can be considered as incoming activations from the $l$-th block, while $\y^{[l+1]}$ and $\x^{[l+1]}$ are the activations computed in the $l+1$-st block.

\subsection{Residual networks}
\label{sec:residual_networks}

Residual networks were originally proposed by \citet{He2016} in the context of image recognition. A generic \emph{residual block} has the specific form
\begin{equation}
  \label{eq:residual_block}
  \relu(\underbrace{\W^{[2]}\relu(\W^{[1]}\x+\b^{[1]}) + \b^{[2]}}_{\approx\mathcal{H}(\x) - \x} + \x),
\end{equation}
where $\x$ is the input vector and $\mathcal{H}(\x)$ is an underlying mapping to be fit by a number of stacked layers. Recall that the only difference between \eqref{eq:residual_block} and a plain two layer block is the addition of $\x$ in the end. This structure is motivated by the degradation problem which typically arises when additional layers are added to a deep network. In theory, the optimal training accuracy of the deeper network cannot be larger compared to the training accuracy of the shallower network due to the fact that the learner can principally choose the additional layer to be an identity map, while all other weights can be copied from the shallower network. However, in practice, training accuracy often degrades rapidly when the depth of the network is increased. Therefore, \citet{He2016} argue that a residual block \eqref{eq:residual_block} can serve as a preconditioner in the sense that an identity map can be obtained simply by setting $\W^{[1], [2]} = \zero$ and $\b^{[1], [2]} = \zero$. Indeed, experiments on various datasets have shown that stacking residual blocks \eqref{eq:residual_block} instead of plain layers leads to increasing training accuracy in practice.

\subsection{Primal-dual residual networks}
\label{sec:primal_dual_residual_networks}

\begin{figure}[t]
  \centering
  \begin{tikzpicture}[yscale= 2.4,xscale=2.4,
    place/.style={circle,
      draw=blue!50,
      fill=blue!20,thick,
      minimum width={1.1cm},},
    transition/.style={rectangle,draw=black!50,fill=black!20,thick}]]

    \node (yl) at (0,0) [circle, draw=black] {$\b^{[l+1]}$};

    \node (xl) at (1.5,1) [place] {$\x^{[l]}$};
    
    \node (xl+1) at (4.5,1) [place] {$\x^{[l+1]}$}; 
    
    \node (+yl+1) at (1.5,0) [transition] {$+$};
    
    \node (+xl+1) at (3,1) [transition] {$+$};+
    
    \node (yl+1) at (3,0) [place] {$\tilde\y^{[l+1]}$};

    \draw[->] (yl) to (+yl+1);
    
    \draw[->] (xl) to (+xl+1);
    
    \draw[->] (xl) to node [right,midway] {$\W^{[l+1]}$} (+yl+1);
    
    \draw[->] (+yl+1) to node [above,midway] {$\prox_{\sigma F^{*}}$} (yl+1);
    
    \draw[->] (yl+1) to node [right,midway] {$\V^{[l+1]}$} (+xl+1);
    
    \draw[->] (+xl+1) to node [above,midway] {$\prox_{\tau G}$} (xl+1);

    \draw[<-] (xl) -- ++(-0.325,0) edge[-,dashed] ++ (-0.325,0);

    \draw (xl+1) -- ++(0.325,0) edge[->, dashed] ++ (0.325,0);
    \draw (xl+1) -- ++(0,-0.325) edge[->, dashed] ++ (0,-0.325);
    
  \end{tikzpicture}
  \caption{Primal-dual residual block without extrapolation.}
  \label{fig:primal_dual_residual_block}
\end{figure}
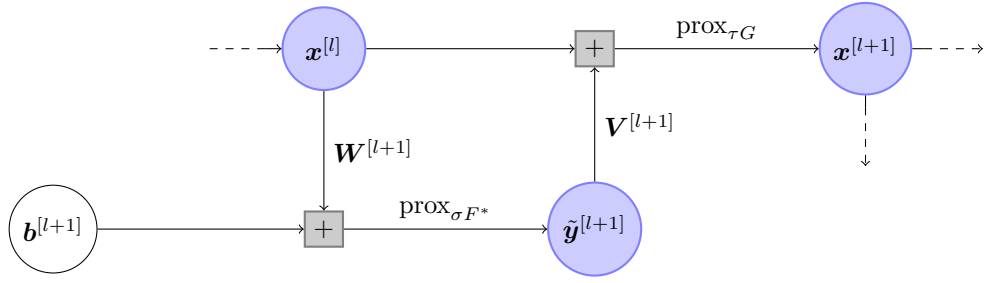

It turns out that the network structure illustrated in Figure \ref{fig:primal_dual_block} manifests two different kinds of such residual blocks. Namely, the activations are computed according to
\begin{align}
  \label{eq:dual_residual_block}
  &\y^{[l+1]} = \prox_{\sigma F^{*}}(\W^{[l+1]}\prox_{\tau G}(\V^{[l]}\y^{[l]} + \x^{[l-1]}) + \y^{[l]})\quad\text{and}\\
  \label{eq:primal_residual_block}
  &\x^{[l+1]} = \prox_{\tau G}(\V^{[l+1]}\prox_{\sigma F^{*}}(\W^{[l+1]}\x^{[l]} + \y^{[l]}) + \x^{[l]}),
\end{align}
respectively. The dual forward map \eqref{eq:dual_residual_block} has the form of a residual block where a previous primal activation $\x^{[l-1]}$ plays the role of a bias unit in the anterior layer. Vice versa, the primal forward map \eqref{eq:primal_residual_block} can be seen as a residual block where a previous dual activation $\y^{[l]}$ acts as bias unit. Thus, in some sense, a primal-dual network encompasses both one residual network with respect to the dual variables and one  with respect to the primal variables, where both networks overlap and interchange information through bias units. Moreover, if we replace the dual activation $\y^{[l]}$ by a proper bias unit $\b^{[l+1]}$ in (\ref{eq:primal_residual_block}), then the resulting mapping has the form of a residual block (\ref{eq:residual_block}), except that the ReLU activation functions are replaced by proximal mappings and that there is only one bias unit. However, the proposed replacement breaks the structure of overlapping networks. Formally, we obtain a \emph{primal-dual residual block}
\begin{equation}
  \label{eq:primal_dual_residual_block}
  \begin{cases}
    \tilde \y^{[l + 1]} &\hspace{-.75em} = \prox_{\sigma F^{*}}(\W^{[l+1]}\x^{[l]}+ \b^{[l+1]}),\\
    \x^{[l+1]} &\hspace{-.75em} = \prox_{\tau G}(\V^{[l+1]}\tilde \y^{[l+1]} + \x^{[l]}),
  \end{cases}
\end{equation}
where $\tilde \y^{[l+1]}$ denotes the activation in the intermediate layer. Figure \ref{fig:primal_dual_residual_block} illustrates a generic primal-dual residual block. Compared to a classical primal-dual block (see Figure \ref{fig:primal_dual_block}), the main difference is that the skip connections between each two dual activations $\y^{[l]}$ and $\y^{[l+1]}$ are cancelled. Instead, the new intermediate activation $\tilde \y^{[l+1]}$ receives a trainable bias unit $\b^{[l+1]}$ as input.

\section{Application to speech dequantization}
\label{sec:speech_dequantization}

In the context of speech dequantization, \cite{Brauer2016} consider the problem
\begin{equation}
  \label{eq:speech_dequantization}
  \a^{\star} \in \argmin_{\a\in\R^{n}} \ \norm{\a}_{1} \quad  \mathrm{s.t.} \ \norm{\K^{-1}\a - \q}_{\infty} \leq \tfrac{\Delta}{2},
\end{equation}
 where $\q\in\R^{n}$ is the uniformly quantized version of a speech signal $\s^{\dagger}\in\R^{n}$ which shall be reconstructed. The scalar $\Delta > 0$ is the length of the quantization intervals and the constraint in (\ref{eq:speech_dequantization}) takes into account that $\s^{\dagger}$ originates in an $\ell_{\infty}$-norm ball with radius $\Delta / 2$ around the quantized signal. Accordingly, $\a^{\star}\in\R^{n}$ approximates $\s^{\dagger}$ in terms of the columns of a dictionary $\K^{-1}\in\R^{n\times n}$ which is assumed to have full rank, and the $\ell_{1}$-norm penalty incorporates the assumption that the sought representation is sparse. Finally, the reconstructed signal is obtained as $\s^{\opt} \coloneqq \K^{-1}\a^{\star}$. Since the dictionary $\K^{-1}$ is by assumption invertible, one can perform the change of variables $\x\coloneqq \K^{-1}\a - \q$ and solve the problem
 \begin{equation}
   \label{eq:speech_dequantization_substituted}
   \x^{\opt} \in \argmin_{\x\in\R^{n}} \ \norm{\K\x + \K\q}_{1} \quad \mathrm{s.t.} \ \norm{\x}_{\infty} \leq \tfrac{\Delta}2
 \end{equation}
 instead. By means of $F(\y) \coloneqq \norm{\y + \K\q}_{1}$ and $G(\x)\coloneqq I_{\norm{\cdot}_{\infty} \leq \frac{\Delta}2}(\x)$, the problem (\ref{eq:speech_dequantization_substituted}) can be written in the form (\ref{eq:primal_problem}) with both functions being proper, convex and lower-semicontinuous. Note that $G$ is the indicator function of the feasible set of (\ref{eq:speech_dequantization_substituted}), i.e., $G(\x) = 0$ if $\norm{\x}_{\infty} \leq \Delta / 2$ and $G(\x) = \infty$ in any other case.

\subsection{Related network architectures}
\label{sec:related_network_architectures}
 
On the one hand, the conjugate function of $F$ also involves an indicator function, namely $F^{*}(\y) = I_{\norm{\cdot}_{\infty}\leq 1}(\y) - (\K\q)^{\top}\y$. On the other hand, the proximal operator of an indicator function of a convex set is simply the projection onto this set. Hence, also taking the linear part in $F^{*}$ into account, we obtain the associated primal-dual block
\begin{equation}
  \label{eq:primal_dual_block_speech}
  \begin{cases}
    \y^{[l+1]} &\hspace{-.75em} = {\cal P}_{B_{1}^{\infty}}(\W^{[l+1]}(\x^{[l]}+\q)+\y^{[l]}),\\
    \x^{[l+1]} &\hspace{-.75em} = {\cal P}_{B_{\Delta / 2}^{\infty}}(\V^{[l+1]}\y^{[l+1]}+\x^{[l]})
  \end{cases}
\end{equation}
according to (\ref{eq:network_block}), where ${\cal P}_{B_{r}^{\infty}}(\x) \coloneqq \max(\min(\x, \, r\one), \, -r\one)$ is the projection onto the $\ell_{\infty}$-norm ball with center $\bm 0$ and radius $r$. In the same way, the primal-dual residual block
\begin{equation}
  \label{eq:primal_dual_residual_block_speech}
  \begin{cases}
    \tilde\y^{[l+1]} &\hspace{-.75em} = {\cal P}_{B_{1}^{\infty}}(\W^{[l+1]}(\x^{[l]}+\q)+\b^{[l+1]}),\\
    \x^{[l+1]} &\hspace{-.75em} = {\cal P}_{B_{\Delta / 2}^{\infty}}(\V^{[l+1]}\tilde\y^{[l+1]}+\x^{[l]})
  \end{cases}
\end{equation}
according to (\ref{eq:primal_dual_residual_block}) can be formed. Note that the addition of $\q$ is due to the term $-(\K\q)^{\top}\y$ which occurs in $F^{*}$ and causes a constant linear offset in the respective proximal operator, namely, $\prox_{\sigma F^{*}}(\y) = {\cal P}_{B_{1}^{\infty}}(\y + \K\q)$. Moreover, the projection operator ${\cal P}_{B_{r}^{\infty}}$ is a generalization of the well-known $\htanh$ activation function \citep{Collobert2004}, i.e., for $x\in\R$ it holds that
\begin{equation}
  \label{eq:hardtanh}
  {\cal P}_{B_{1}^{\infty}}(x) = \max(\min(x, \, 1), \, -1) = \htanh(x).
\end{equation}
For $\x\in\R^{n}$, the projection onto the ball with radius $r$ can be obtained by componentwise application of a scaled $\htanh$ function, namely ${\cal P}_{B_{r}^{\infty}}(\x)_{j} = \max(\min(x_{j}, \, r), \, -r) = r\cdot\htanh(r^{-1}x_{j})$.


\section{Numerical Experiments}
\label{sec:numerical_experiments}

We investigate the impact of primal-dual (residual) networks using a dataset of 720 sentences from the IEEE corpus provided in \cite{Loizou2013} consisting of male speech and sampled at 16 kHz. From these signals, 70\% (i.e., 504 signals) are used as training set, 15\% (i.e., 108 signals) are reserved as development set, and another 15\% serve as test set which we use for a comparison with the results of \cite{Brauer2016}. In order to reconstruct ground truth signals $\s^{\dagger}$ from quantized measurements $\q$, our goal is to train primal-dual (residual) networks consisting of $L$ stacked blocks \eqref{eq:primal_dual_block_speech} and \eqref{eq:primal_dual_residual_block_speech}, respectively. As initial activations, we use $\x^{[0]} = \zero$ and $\y^{[0]}= \tilde\y^{[0]}= \zero$. However, as discussed above, the actual input of the network $\q$ appears in the form of a linear offset during the computation of each dual activation. Our estimate for $\s^{\dagger}$ is finally $\hat\s \coloneqq h_{\Theta}(\q)\coloneqq \x^{[L]} + \q$, where $\Theta$ denotes the aggregate of weights in the network.

\subsection{Learning setup}
\label{sec:learning_setup}

To make the networks $h_{\Theta}$ principally usable for real-time applications, the original and quantized signals are truncated using a rectangular window function. To that end, a window size $n$ as well as a shift length $s\leq n$ are a priori fixed. Then, the signals are split into $n$-dimensional sub-signals
\begin{equation}
  \label{eq:signal_splitting}
  \s^{(j)}\coloneqq (\s^{\dagger}_{j s + 1},\dots,\s^{\dagger}_{j s + n}) \quad \text{and} \quad \q^{(j)} \coloneqq (\q_{j s + 1},\dots,\q_{j s + n}), \quad j = 0,\dots,\left\lfloor\tfrac{N - n}{s}\right\rfloor,
\end{equation}
where $N$ is the dimension of the respective full signals. We use $n = s = 1024$ and truncate all 504 signals in the training set. This way, we end up with $m = 20641$ training examples and a new training set $\{(\q^{(i)}, \s^{(i)}) : i = 1,\dots,m\} \subset \R^{n}\times\R^{n}$. To train the network, we use a weighted sum of mean squared error (MSE) and regularization as loss function, i.e., we minimize
\begin{equation}
  \label{eq:loss}
  \mathcal{L}(\Theta) \coloneqq \frac{1}{mn}\sum_{i = 1}^{m}\norm{h_{\Theta}(\q^{(i)}) - \s^{(i)}}_{2}^{2} + \lambda{\cal R}(\Theta).
\end{equation}
Subsequently, we use $\ell_{2}$-regularization and choose the hyperparameter $\lambda \geq 0$ with respect to the development set. All experiments were conducted on an NVIDIA GeForce\textsuperscript{\textregistered} GTX 1080 Ti GPU using TensorFlow\texttrademark 1.5.0. To minimize (\ref{eq:loss}), we used Adam \citep{Kingma2014} with learning rate $10^{-4}$ and all other parameters set to standard values.

\subsection{Comparison of network architectures}
\label{sec:comparison_resnets}

\begin{table}[t]
  \centering
  \caption{Minimum MSE values (in multiples of $10^{-4}$) achieved with PDNs and PDRNs}
  
  \begin{tabular}{llllllll}
    \toprule
    &&&\multicolumn{5}{c}{$L$}\\
    \cmidrule(r){4-8}
    &data &reg. &1 &2 &5 &10 &15\\
    \midrule
    PDN &train. &w/o &2.08 &1.28 &0.48 &0.28 &1.51\\
    PDRN &train. &w/o &2.05 &1.28 &0.53 &0.40 &0.57\\
    PDN &train. &w/ &3.29 &2.96 &2.06 &1.62 &1.69\\
    PDRN &train. &w/ &2.09 &1.57 &1.06 &1.01 &1.06\\
    \midrule
    PDN &dev. &w/o &2.43 &2.36 &2.31 &2.37 &2.48\\
    PDRN &dev. &w/o &2.38 &2.32 &2.00 &1.92 &2.03\\
    PDN &dev. &w/ &3.46 &3.20 &2.49 &\textbf{2.07} &2.13\\
    PDRN &dev. &w/ &2.40 &2.24 &1.84 &\textbf{1.81} &1.88\\
    \bottomrule
  \end{tabular}
  \label{tab:comparison_of_network_architectures}
\end{table}

\begin{figure}
  \centering

  \includegraphics[width=.49\textwidth]{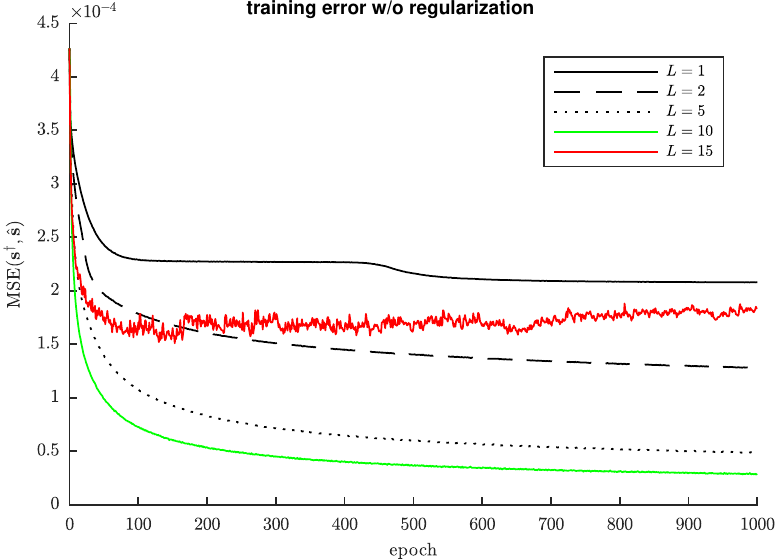}
  \includegraphics[width=.49\textwidth]{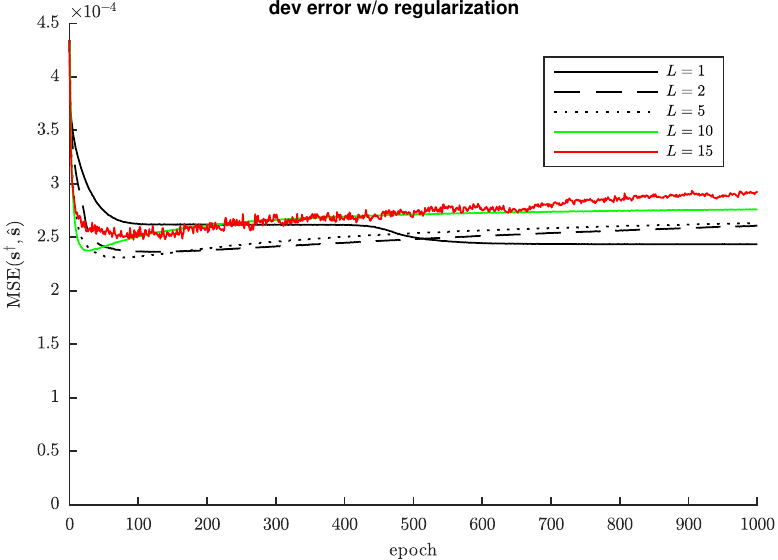}\\
  \medskip
  
  \includegraphics[width=.49\textwidth]{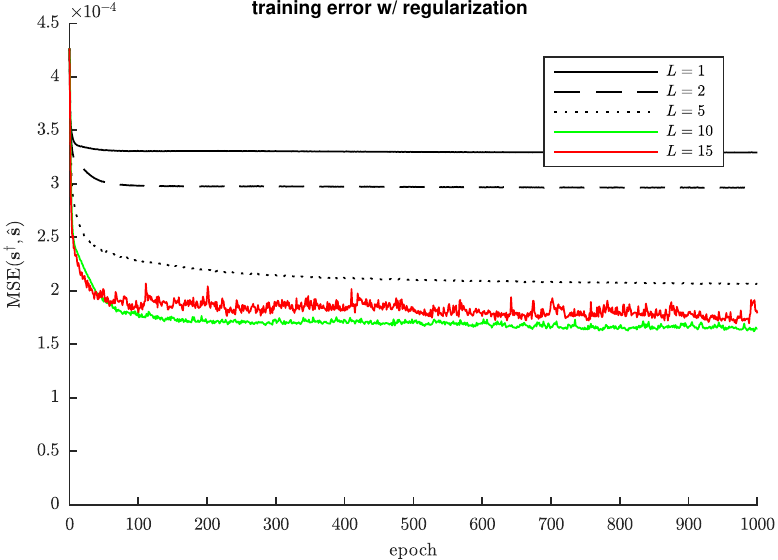}
  \includegraphics[width=.49\textwidth]{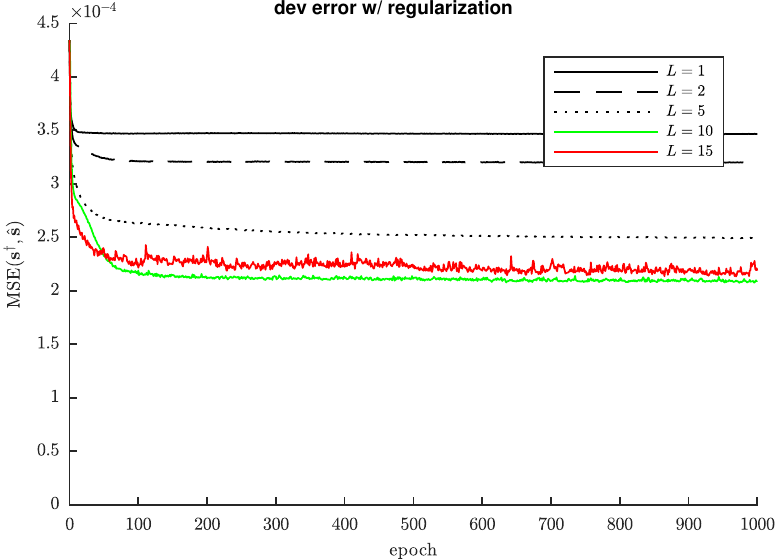}
  
  \caption{Training and development errors yielded with PDNs (\ref{eq:primal_dual_block_speech}).}
  \label{fig:results_pd}
\end{figure}

\begin{figure}
  \centering

  \includegraphics[width=.49\textwidth]{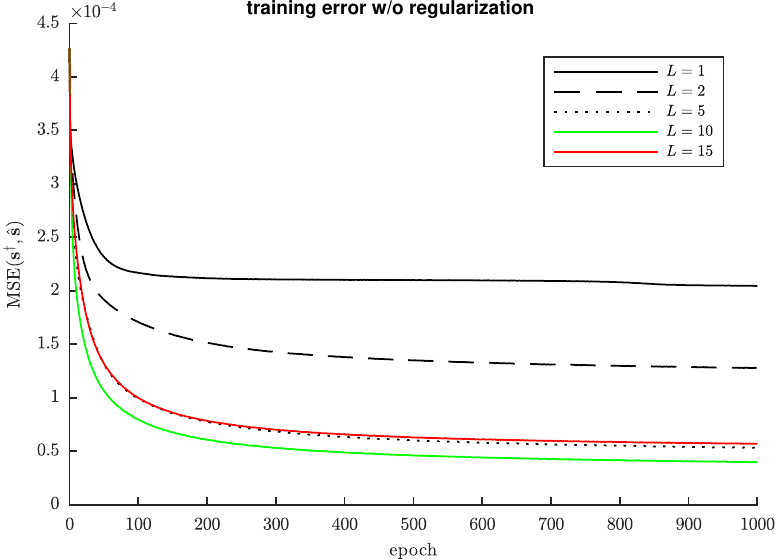}
  \includegraphics[width=.49\textwidth]{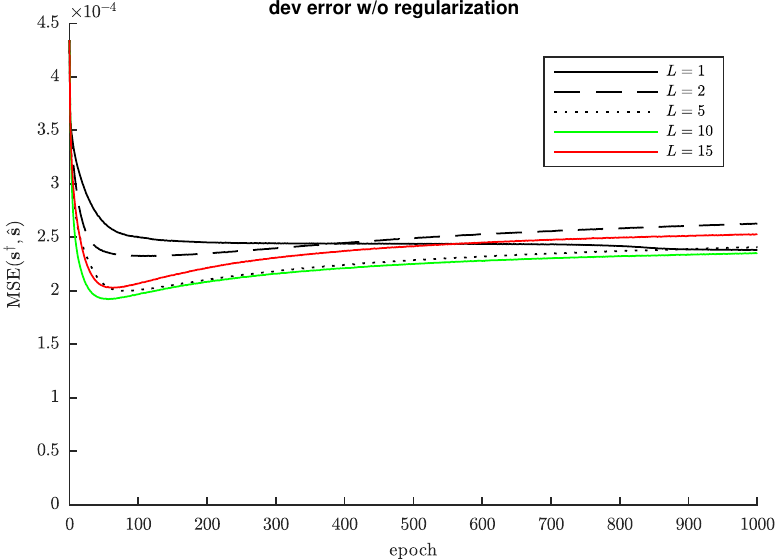}\\
  \medskip
  
  \includegraphics[width=.49\textwidth]{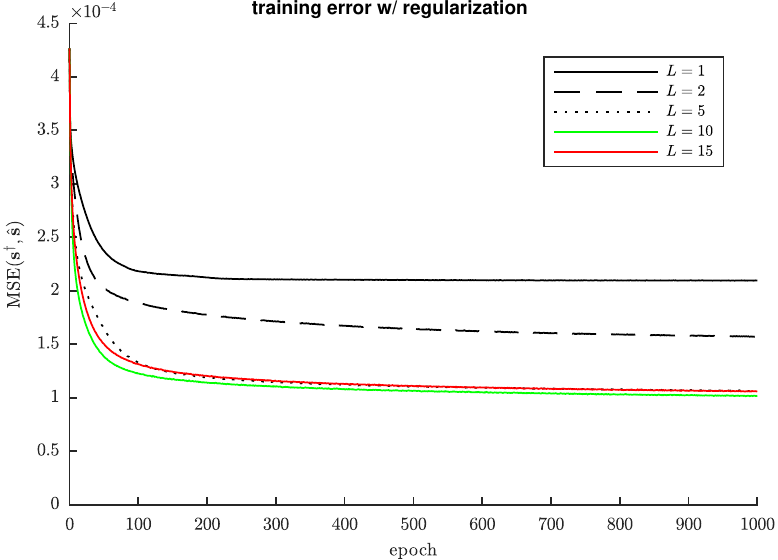}
  \includegraphics[width=.49\textwidth]{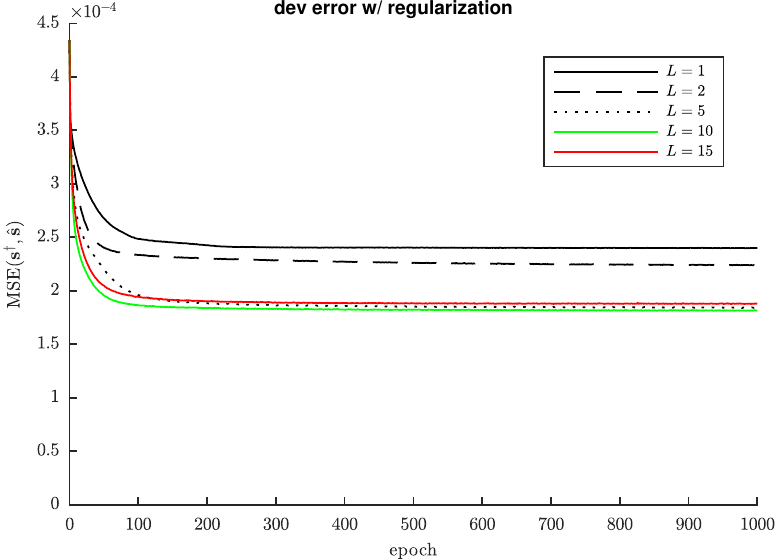}
  \caption{Training and development errors yielded with PDRNs (\ref{eq:primal_dual_residual_block_speech}).}
  \label{fig:results_pd_res}
\end{figure}

In a first set of experiments, we compared the impact of primal-dual networks (PDNs) and primal-dual residual networks (PDRNs) applied to the task of speech dequantization. In either case, we tried different depths $L\in\{1, 2, 5, 10, 15\}$ and trained each network over 1000 epochs using a batch size of 128. Our results are illustrated in Figures \ref{fig:results_pd} and \ref{fig:results_pd_res}, and in Table \ref{tab:comparison_of_network_architectures} we report the minimum MSE values associated with each trained network.

All in all, our results indicate that PDRNs feature superior performance compared to classical PDNs. On the one hand, Table \ref{tab:comparison_of_network_architectures} shows that PDRNs yield throughout lower MSE values relative to PDNs on the development set (17\% lower without regularization and 13\% lower with regularization, if one considers the respective best values among all depths). On the other hand, it can be seen clearly from Figures \ref{fig:results_pd} and \ref{fig:results_pd_res} that, especially in case of deeper networks, PDRNs exhibit a more stable behavior in the sense that training and development errors decay smoother over time. In this context, the displayed results for $L = 15$ are particularly notable. In addition, we could observe similar behavior for depths $L > 15$.

\subsection{Comparison with primal-dual algorithm}
\label{sec:comparison_primal_dual}

In further experiments conducted on the test set, we compared PDNs and PDRNs to the Chambolle-Pock algorithm (CP). To that end, we computed the respective MSEs as above and, in addition, the associated signal-to-noise ratios (SNRs). On the side of neural networks, we selected both one PDN and one PDRN on the basis of the respective errors on the development set (cf. the bold-faced entries in Table \ref{tab:comparison_of_network_architectures}). Further, we applied CP to problem (\ref{eq:speech_dequantization}) with $K = \mathrm{DCT}$ being a discrete cosine transform matrix, exactly as described in \cite{Brauer2016}. For the benefit of a fair comparison, we stopped CP after 10, 25 and 50 iterations and calculated the respective errors at that points. Supplementary, we computed the MSEs as well as the SNRs of the quantized signals before reconstruction (QU). Our results are reported in Table \ref{tab:comparison_with_primal_dual_algorithm}.

A comparison of the test results in Table \ref{tab:comparison_with_primal_dual_algorithm} reveals that PDNs and PDRNs offer significantly lower MSE values and higher SNR values than the Chambolle-Pock algorithm, seemingly independent of the performed number of iterations. The considered PDRN improves the MSE of the unquantized signal by 58.6\% and its SNR by 25.5\%, whereas the largest number of Chambolle-Pock iterations yields improvements of only 32.9\% and 11.9\%, respectively. Further, it can be seen that PDRNs outperform PDNs once more, since the respective relative improvements yielded with the considered PDN amount to 52.3\% and 21.3\%.

\begin{table}[b]
  \centering
  \caption{Test errors achieved with quantized signals before reconstruction (QU), best PDN and PDRN w.r.t. development set, and Chambolle-Pock algorithm (CP) (MSE values in multiples of $10^{-4}$)}
  
  \begin{tabular}{rrrrrrr}
    \toprule
    &&&&\multicolumn{3}{c}{\# CP iterations}\\
    \cmidrule(r){5-7}
    &QU &PDN &PDRN &10 &25 &50\\
    \midrule
    MSE &4.47 &2.13 &1.85 &3.05 &3.00 &3.00\\
    SNR &15.06 &18.27 &18.90 &16.71 &16.81 &16.85\\
    \bottomrule
  \end{tabular}
  \label{tab:comparison_with_primal_dual_algorithm}
\end{table}

\section{Conclusion}
\label{sec:conclusion}

In this paper, we have proposed the architecture of primal-dual residual networks which combines features of unrolled proximal splitting methods and residual networks. Further, we have drawn a comparison between the proposed network architecture and classical primal-dual networks which can be considered straightforwardly unrolled proximal splitting methods. Our results have shown that, applied to speech dequantization, primal-dual residual networks can outperform their classical counterpart significantly. Moreover, we have seen that both architectures can beat a truncated proximal splitting scheme on the same task.

However, it should not go unmentioned that some questions have been left open for future research. First, we have yet only compared primal dual (residual) networks without extrapolation. As extrapolation is an important aspect concerning the convergence of proximal splitting methods, it may also play a role in view of the behavior of the related network architectures. Second, using the example of speech dequantization, we have shown how a particular convex optimization problem with hard constraints can be unrolled in terms of a neural network by performing an appropriate change of variables and considering the constraint as an indicator function of a related norm ball. It is likely possible that this approach can be generalized to a larger class of optimization problems. We are going to address these as well as other related questions in the near future.

\bibliographystyle{plainnat}
\bibliography{references}

\end{document}